\documentclass[11pt]{article}
\usepackage{eamt26}
\usepackage{times}
\usepackage{url}
\usepackage{latexsym}
\usepackage[small,bf]{caption} 
\usepackage{colortbl,xcolor}
\usepackage{multirow}
\usepackage{hyperref}
\usepackage{placeins}

\hypersetup{
    colorlinks=true,
    linkcolor=[rgb]{0,0,0.6},   
    urlcolor=[rgb]{0,0,0.6},
    citecolor=[rgb]{0,0,0.6},
    pdfborder={0 0 0}           
}

\usepackage{lipsum}
\usepackage{listings}

\usepackage[utf8]{inputenc}
\usepackage[T1]{fontenc}
\usepackage{graphicx}  
\usepackage{ifthen}
\newboolean{reviewmode}
\setboolean{reviewmode}{false}  

\title{CompactQE: Interpretable Translation Quality Estimation\\ via Small Open-Weight LLMs}

\ifthenelse{\boolean{reviewmode}}{
    \author{Anonymous submission}
}{
    \author{
    Kamil Guttmann$\normalfont{\textsuperscript{1,2}}$, 
    Zofia Fraś$\normalfont{\textsuperscript{1}}$, 
    Artur Nowakowski$\normalfont{\textsuperscript{1,2}}$, 
    Krzysztof Jassem$\normalfont{\textsuperscript{1,2}}$ \\
    $\textsuperscript{1}$ Laniqo, Poznań, Poland \\
    $\textsuperscript{2}$ Faculty of Mathematics and Computer Science, Adam Mickiewicz University, Poznań, Poland \\
    {\tt \{name\}.\{surname\}@laniqo.com}
    }
}

\date{}

\begin{document}
\maketitle
\begin{abstract}
Current state-of-the-art Quality Estimation (QE) in machine translation relies on massive, proprietary LLMs, raising data privacy concerns. We demonstrate that smaller, open-source LLMs (<30B parameters) are a viable, cost-effective and privacy-preserving alternative. Using a single-pass prompting strategy, our models simultaneously generate quality scores, MQM error annotations, suggested error corrections, and full post-editions. Our analysis shows these models achieve highly competitive system-level correlations with human judgments that outperform traditional neural metrics, fine-tuned models, and human inter-annotator agreement, effectively approximating the capabilities of much larger proprietary LLMs.
\end{abstract}

\section{Introduction}

Quality Estimation (QE) remains a critical component in professional machine translation workflows, enabling the assessment of translation reliability without access to reference translations. While neural regression-based metrics such as COMET \cite{rei-etal-2020-comet} have achieved high correlations with human judgments, they typically produce a single scalar score. This aggregation provides little insight into specific translation errors or their severity, limiting the utility of such metrics for downstream tasks that require detailed feedback and interpretability. Consequently, there is a growing demand for evaluation methods that can offer fine-grained error diagnosis alongside quality scores.

The emergence of Large Language Models (LLMs) has introduced a new paradigm for evaluation, often referred to as "LLM-as-a-judge". Approaches such as GEMBA \cite{kocmi-federmann-2023-large} and AutoMQM \cite{fernandes-etal-2023-devil} leverage the generative capabilities of LLMs to mimic human evaluation frameworks like Multidimensional Quality Metrics (MQM) \cite{burchardt-2013-multidimensional} or Error Span Annotation (ESA) \cite{kocmi-etal-2024-error}. By employing zero-shot or few-shot prompting strategies, these methods can identify and categorize error spans, offering a level of interpretability that was previously difficult to attain with standard neural metrics. 

However, the state-of-the-art performance of these generative metrics currently relies heavily on massive, proprietary models such as GPT-5 or Gemini. This dependency presents significant challenges for professional deployment, particularly concerning data privacy. Sending sensitive content to external APIs for evaluation is often prohibitive for enterprises with strict data security requirements. Furthermore, the lack of transparency regarding the training data and architecture of closed models raises concerns about reproducibility and the stability of evaluation standards over time.

We hypothesize that high-quality, interpretable QE can be achieved without reliance on very large proprietary models. In this work, we demonstrate the feasibility of developing an effective MQM-style metric using significantly smaller, open-source multilingual models. By shifting to open weights, we address the critical need for privacy-preserving evaluation pipelines that can be deployed locally within secure environments. We show that with appropriate prompting and output processing, these smaller models can offer more efficient alternatives to their larger, closed-source counterparts.

To achieve competitive performance with these resource-constrained architectures, we introduce a streamlined methodology that integrates quality estimation and post-editing. Building upon the structured prompting strategies of GEMBA-MQM V2 \cite{junczys-dowmunt-2025-gemba}, our approach performs translation post-editing and error detection in a single inference step. This contrasts with recent multi-stage approaches like MQM-APE \cite{lu-etal-2025-mqm}, which require multiple model calls to verify errors. We subsequently apply heuristic error filtering to refine the output, reducing computational overhead while maintaining the granularity required for professional quality checks.

\section{Related Work}

The COMET-Kiwi \cite{rei-etal-2022-cometkiwi} family of metrics and recent models like MetricX-25-QE \cite{juraska-etal-2025-metricx} represent state-of-the-art neural regression baselines explicitly fine-tuned for reference-free translation quality estimation. These models leverage pretrained multilingual encoders to predict a single quality score. While the scalar score can be a good indicator of the translation's quality, it is not easily interpretable.

xCOMET \cite{guerreiro-etal-2024-xcomet} bridges the gap between regression-based scoring and fine-grained error detection by integrating both tasks into a single learned metric. It predicts sentence-level scores while simultaneously identifying error spans and assigning them severity labels (Minor, Major, Critical), achieving high correlation with human scores. While it is a big leap from scalar metrics it still lacks the fine-grained error categorization found in human-based error annotations such as MQM.

Generative, LLM-based approaches like Error Analysis Prompting (EAPrompt) \cite{lu-etal-2024-error} attempt to emulate the human MQM framework by combining Chain-of-Thought (CoT) reasoning with structured error analysis. EAPrompt explicitly instructs the model to identify major and minor errors within its reasoning steps before calculating a final score. While the internal CoT process identifies specific errors, the primary output focus remains the aggregated score or error count, rather than returning a structured list of error spans to the user. Furthermore, EAPrompt employs multi-turn prompting, separating error identification from error counting, which increases the computational cost by requiring multiple model calls per evaluated segment.

AutoMQM represents one of the initial efforts to fully automate the annotation of translations with MQM-style errors using Large Language Models. By leveraging large models like PaLM-2, AutoMQM demonstrated that LLMs could effectively identify and categorize errors when provided with few-shot examples sampled from human annotations. This approach shifted the focus from assigning a scalar score to generating structured error annotations.

The GEMBA family of metrics established the state-of-the-art for LLM-based evaluation using zero-shot and few-shot prompting with GPT models. GEMBA-MQM \cite{kocmi-federmann-2023-gemba} specifically prompts GPT-4 to output structured error annotations, mapping severities to a final quality score. The most recent iteration, GEMBA-MQM V2, addresses the stochastic nature of LLM outputs by aggregating scores across ten distinct inference runs. While this aggregation strategy significantly improves reliability and correlation with human judgments, it increases computational demands, requiring large proprietary models and substantially higher inference time and cost per segment.

MQM-APE builds upon the generative evaluation paradigm by integrating an Automatic Post-Editing (APE) step to verify the impact of detected errors. It filters out non-impactful errors by checking if fixing them actually improves the translation quality according to a pairwise verifier. Notably, MQM-APE demonstrated the viability of using open-source models like Llama-3 \cite{grattafiori2024llama3herdmodels} and Mixtral \cite{jiang2024mixtralexperts} for this task. However, its architecture relies on a sequential pipeline of three distinct modules: evaluator, post-editor, and verifier. Since all of these modules are operated by the LLM, it results in high inference costs.

Our work builds directly upon the structured prompting strategies of GEMBA-MQM V2 but adapts them for resource-constrained environments. We extend the framework by incorporating simultaneous post-editing and ESA scoring within a single inference step, alongside more detailed error descriptions. Unlike GEMBA-MQM V2, which utilizes full document context, we omit document-level information to accommodate the context window limitations of smaller open-source models deployed on consumer-grade GPUs (up to 32GB of VRAM). This trade-off allows us to maintain high interpretability and accuracy while ensuring data privacy and reducing computational overhead.

\section{Methodology}

\subsection{Data}

For all experiments we utilize the official data and human annotations provided by the organizers of the WMT25 Metrics Shared Task. The dataset comprises paragraph-level translations generated by various systems submitted to the WMT25 General MT Task. Each of the system outputs was originally evaluated by the task organizers or participants using multiple automated metrics, including COMET22, COMETKiwi22, MetricX-25-QE, GEMBA-v2 and Gemini2.5-Pro, which serve as the baselines for our proposed approach. Subsequently, the organizers selected a subset of the best scoring systems for human evaluation.

Human evaluation followed the ESA protocol, which has been established as the standard evaluation framework at WMT for the past two iterations. According to the ESA guidelines, human evaluators assess translation quality through a two-step process. First, the annotators identify and highlight specific error spans within the translated text, categorizing them by severity. Second, after completing the span annotation, the evaluators assign an overall quality score ranging from 0 to 100 to the translation. Each segment was evaluated by two annotators. We take the average of the two scores and treat it as the golden standard.

From the complete WMT25 dataset, we select three language pairs to evaluate our method across diverse linguistic scenarios: Czech-German (to represent a non-English-centric translation direction), English-Italian (representing a high-resource language pair), and English-Ukrainian (representing a low-resource target language utilizing a non-Latin script). To ensure a fair and consistent comparison between human judgments and automated metrics, we restrict our experiments strictly to the segments that possess complete human ESA annotations and scores. The resulting number of systems and segments per system is reported in Table~\ref{tab:data}. 

\begin{table}[t]
\centering
\begin{tabular}{|p{0.6\columnwidth}|p{0.3\columnwidth}|}
\hline
\multicolumn{2}{|c|}{\cellcolor{gray!20}\textbf{\small Czech → German}} \\
\hline
Number of systems & 20 \\
Segments per system & 231 \\[0.1em]
\hline
\multicolumn{2}{|c|}{\cellcolor{gray!20}\textbf{\small English → Italian}} \\
\hline
Number of systems & 18 \\
Segments per system & 215 \\[0.2em]
\hline
\multicolumn{2}{|c|}{\cellcolor{gray!20}\textbf{\small English → Ukrainian}} \\
\hline
Number of systems & 18 \\
Segments per system & 199 \\[0.2em]
\hline
\end{tabular}
\caption{Details on the number of systems and segments per language pair. Each segment consisted of multiple sentences, making the evaluation paragraph-level.}
\label{tab:data}
\end{table}

\subsection{Prompts}

To query the models for QE, we base our initial prompt structure on GEMBA-MQM V2. This method was among the best-performing metrics in the WMT25 Metrics Task and provides an effective baseline framework for extracting structured error span annotations, with both input and output formatted as JSON.

In its original formulation, the GEMBA-MQM V2 prompt includes the context of the entire document to score individual segments. While proprietary models generally handle large inputs without issues, many small open-source models possess limited context windows. Furthermore, expanding the active context length significantly increases VRAM consumption. This is often impractical for on-premise deployments reliant on consumer-grade hardware due to the associated infrastructure costs and reduced inference speeds. Consequently, we omit the document-level context in our approach, evaluating each paragraph in isolation to fit these hardware constraints.

To calibrate the model's sensitivity and encourage correct JSON output formatting, we provide in-context learning examples within the prompt. By supplying few-shot examples of segments containing high, medium, and zero error counts, we aim to mitigate common issues such as over-annotation (hallucinating non-existent errors) and under-annotation (missing clear mistakes). The examples were specifically selected from translations across multiple language pairs (English-German, Polish-Spanish, French-Italian) to prevent the model from focusing on pair-specific features. This results in a language-pair-independent prompt, which can be applied to any translation pair without the need to modify or replace examples for subsequent experiments.

We modify the prompt to explicitly request the extraction of both source and target error phrases. Capturing the exact target phrase allows for error span annotation within the translated text. Simultaneously, extracting the corresponding source phrase is necessary for identifying omission errors, where a specific phrase present in the source is entirely absent from the translation.

Alongside the error span annotations, the prompt instructs the model to generate a suggested correction for each identified mistake. Providing targeted fixes has practical utility in professional translation workflows. It can accelerate the human post-editing process by allowing translators to review and accept specific suggestions directly in their computer-assisted translation (CAT) tools.

Finally, we ask the model to output a full post-edited version of the target segment within the same prompt. By consolidating QE, error span annotation and automatic post-edition into a single prompt, we aim to significantly reduce latency and computational overhead.

We provide the system and user prompt templates used for all our experiments in Appendix \ref{appendix:prompt}, Listings \ref{lst:system-prompt}, and \ref{lst:user-prompt}.

\subsection{Models}

We evaluate our approach using three open-source models: \texttt{gemma-3-27b-it}\footnote{\url{https://huggingface.co/google/gemma-3-27b-it}} \cite{gemmateam2025gemma3technicalreport}, \texttt{EuroLLM-9B-Instruct}\footnote{\url{https://huggingface.co/utter-project/EuroLLM-9B-Instruct}} \cite{martins2025eurollm9btechnicalreport}, and \texttt{Qwen3-VL-30B-A3B-Instruct}\footnote{\url{https://huggingface.co/Qwen/Qwen3-VL-30B-A3B-Instruct}} \cite{bai2025qwen3vltechnicalreport}. We selected these specific models because they fall under the 30-billion parameter threshold, enabling on-premise deployment on smaller, consumer-grade GPUs. Furthermore, they are available under non-restrictive open-source licenses and possess strong multilingual capabilities. Although \texttt{Qwen3-VL-30B-A3B-Instruct} is inherently a Vision-Language (VL) Model, we utilize only its text-processing modalities for this task, as it is reported to achieve superior results on text-only benchmarks compared to its non-VL counterpart \cite{bai2025qwen3vltechnicalreport}.

To benchmark our results against state-of-the-art proprietary systems, we additionally run quality estimation using \texttt{Gemini-3-Flash} \cite{gemini3flash2025} as a backbone model. All models are prompted using an identical prompt structure. 

For the open-source models, we apply greedy decoding by setting the temperature to zero (\texttt{T = 0}). Conversely, we run \texttt{Gemini-3-Flash} with \texttt{T = 1}, as decreasing the temperature is not recommended in its documentation. To ensure a fair comparison, we minimize the impact of Gemini's internal reasoning capabilities by setting its thinking budget to zero. While the Gemini 3 family does not support a complete 'thinking-off' mode, this configuration triggers a 'minimal' thinking level, effectively preventing the model from complex reasoning.

\subsection{Error Filtering}

\begin{table*}
\centering
\begin{tabular}{|l|cc|c|} \hline
& \multicolumn{2}{c|}{Error count} & \multirow{2}{*}{Rejection rate} \\
& Before filtering & After filtering & \\ \hline
\multicolumn{4}{|c|}{\cellcolor{gray!20}\textbf{\small Czech → German}} \\ \hline
EuroLLM-9B-Instruct & 20,278 & 13,591 & 33.0\% \\
Qwen3-VL-30B-A3B-Instruct & 44,099 & 21,871 & 50.4\% \\
Gemma-3-27b-it & 12,467 & 10,234 & 17.9\% \\
Gemini-3-Flash & 16,247 & 15,134 & 6.9\% \\ \hline
\multicolumn{4}{|c|}{\cellcolor{gray!20}\textbf{\small English → Italian}} \\ \hline
EuroLLM-9B-Instruct & 11,387 & 7,953 & 30.2\% \\
Qwen3-VL-30B-A3B-Instruct & 30,119 & 16,530 & 45.1\% \\
Gemma-3-27b-it & 8,407 & 7,351 & 12.6\% \\
Gemini-3-Flash & 13,519 & 13,032 & 3.6\% \\ \hline
\multicolumn{4}{|c|}{\cellcolor{gray!20}\textbf{\small English → Ukrainian}} \\ \hline
EuroLLM-9B-Instruct & 12,212 & 8,180 & 33.0\% \\
Qwen3-VL-30B-A3B-Instruct & 24,277 & 15,357 & 36.7\% \\
Gemma-3-27b-it & 9,509 & 7,996 & 15.9\% \\
Gemini-3-Flash & 15,324 & 14,609 & 4.7\% \\ \hline
\end{tabular}
\caption{Total number of errors detected by four QE models across three language pairs, before and after filtering, with rejection rates.}
\label{tab:errors}
\end{table*}

The raw output from generative models often contains inconsistencies, necessitating a structured filtering pipeline. To ensure robust parsing of the generated responses, we utilize the \texttt{json-repair} library\footnote{\url{https://github.com/mangiucugna/json_repair}} \cite{Baccianella_JSON_Repair_-_2025} to automatically correct any malformed JSON structures before further processing. 

Our first filtering step addresses error span hallucinations, where the model identifies a target phrase that does not actually exist in the generated translation. We filter out these errors based on exact string matching after lowercase normalization. 

Furthermore, we observe that models frequently duplicate the same error across multiple severity categories. In these instances, we deduplicate the annotations by retaining only the instance with the highest severity. This heuristic ensures that critical or major translation flaws are accurately represented and not overshadowed or mistakenly downgraded in favor of redundant minor complaints. The filtering results for each model are presented in Table \ref{tab:errors}.

Additionally, we explored a secondary filtering mechanism based on the generated post-editions. This approach used fuzzy matching to compare the model’s suggested correction with its post-edited segment. The underlying logic assumes that if the model flags an error and proposes a fix, but fails to incorporate that fix into its final post-edition, the flagged error is likely a false positive. However, our preliminary tests yielded a low level of improvement when applying this filter. We conclude that this heuristic relies heavily on the model possessing exceptionally strong and consistent post-editing capabilities.

This observation aligns closely with the findings from the WMT25 Metrics Task 3, which focused on post-editing sentences with pre-annotated error spans. Lavie et al.~\shortcite{lavie-etal-2025-findings} noted that while LLMs demonstrate an ability to improve traditional Neural Machine Translation (NMT) outputs, they struggle to meaningfully improve outputs already generated by LLMs. Consequently, we do not include this filtering step, as the current post-editing limitations of smaller open-source models can introduce noise into the filtering process rather than refine it.

\section{Evaluation}

\begin{figure*}[ht]
    \includegraphics[width=\textwidth, trim={0 0cm 0 2.2cm}, clip]{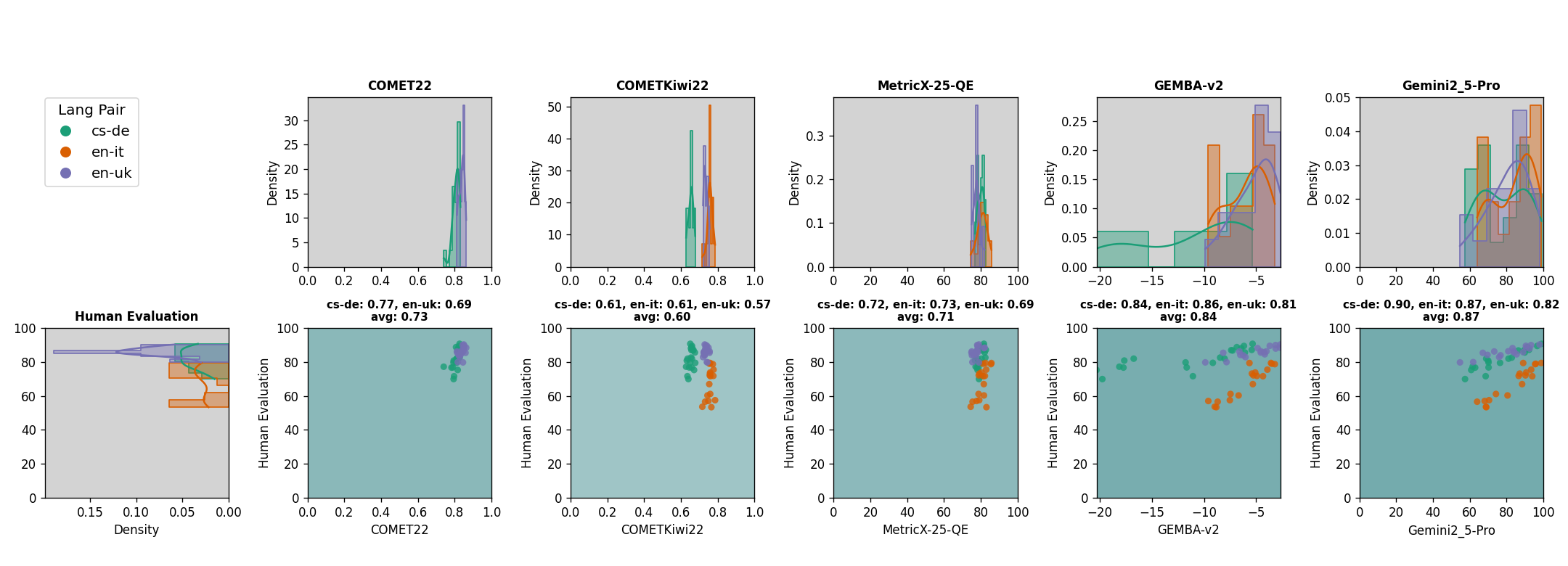}
    \caption{System-level performance of baseline metrics on the WMT25 dataset. The histograms display the score distributions for each metric, while the scatter plots illustrate the correlation between automated scores and human evaluations. Soft Pairwise Accuracy values for each language pair and their respective averages are provided above each plot. The background heatmap indicates the average accuracy across all language pairs (note: EN–IT is excluded for COMET22 due to the lack of available reference translations).}
    \label{fig:pearson_metrics}
\end{figure*}

\begin{figure*}[ht]
    \begin{center}
        \includegraphics[scale=0.35, trim={0 0cm 0 1.5cm}, clip]{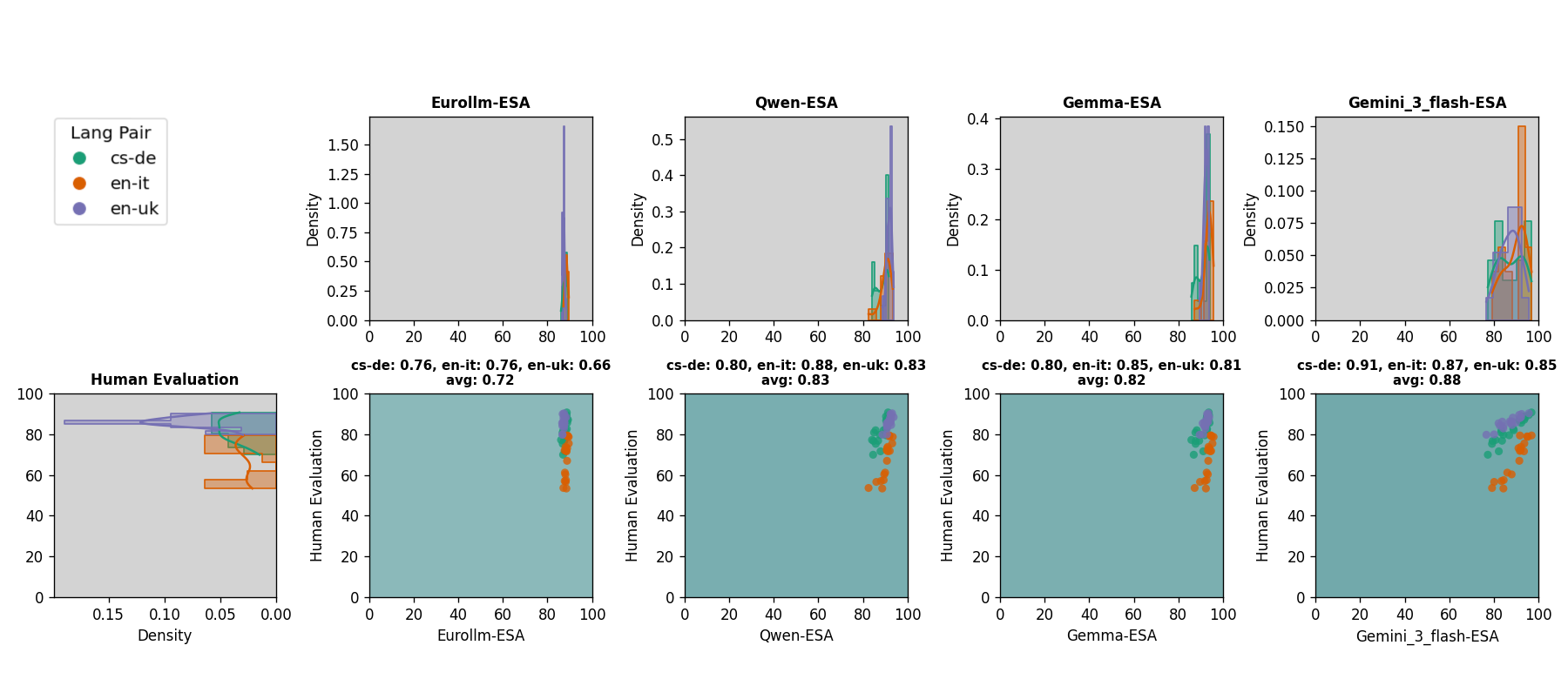}
        \caption{System-level performance of the proposed LLM-based QE metrics. The histograms display the distribution of scores generated by each model, while the scatter plots demonstrate the correlation with human-assigned ratings. Soft Pairwise Accuracy values for each language pair and their overall averages are indicated above each plot. The background heatmap represents the mean accuracy across all language pairs, with all proposed models providing quality scores on a scale of 0 to 100.}
        \label{fig:pearson_qe}
    \end{center}
\end{figure*}

Following the evaluation framework implemented in the \texttt{mt-metrics-eval}\footnote{\url{https://github.com/google-research/mt-metrics-eval}} toolkit, we compute correlations between our QE scores, selected baseline metrics, and human ESA ratings. For the system-level evaluation, we report Soft Pairwise Accuracy (SPA) \cite{thompson-etal-2024-improving}. At the segment level, we evaluate the metrics using the "group-by-item" segment-level accuracy with tie calibration \cite{deutsch-etal-2023-ties}.

To evaluate the accuracy of the generated error spans, we adopt the official protocol established in the WMT25 Metrics Task 2 \cite{lavie-etal-2025-findings}. We calculate the precision, recall, and micro-F1 scores at the character level to measure the overlap between the predicted and gold human error spans. This evaluation scheme rewards not only the exact character matches but additionally it incorporates a partial credit weighting of 0.5 points for cross-severity matches, meaning the model receives partial recognition if it correctly identifies the location of an error but misclassifies its severity.

It is important to note a limitation regarding the alignment of source-side errors within this specific evaluation framework. It accounts only for the errors that can be explicitly mapped to character indices within the target translation string. Consequently, it does not evaluate errors that can be annotated solely in the source sentence. A primary example is an omission error, where a phrase or an entire sentence is left untranslated and thus has no corresponding text span in the target output. For this reason, although our models are prompted to detect and extract these source-side omissions, we ignore them when calculating the F1 score. On the other hand, the omission errors are taken into account by the model when assigning the segment-level scalar score.

\section{Results}

\begin{figure*}[ht]
    \includegraphics[width=\textwidth, trim={0 0cm 0 3.8cm}, clip]{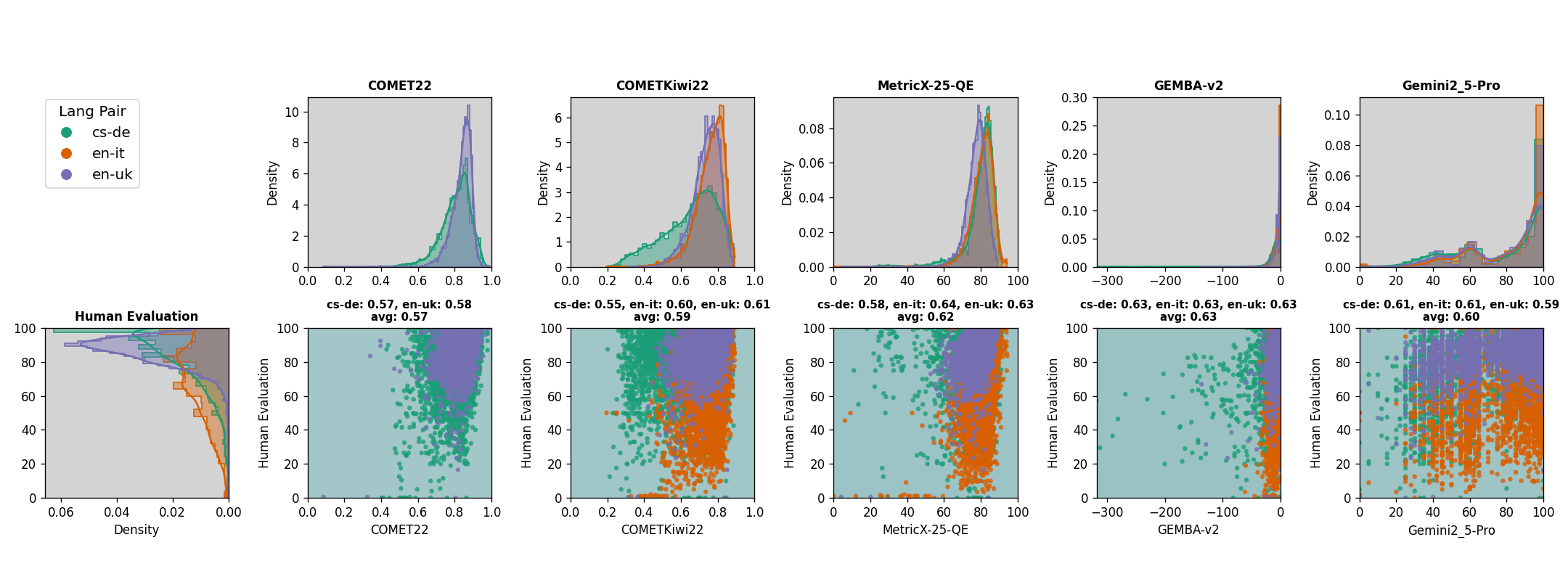}
    \caption{Segment-level performance of baseline metrics on the WMT25 dataset. The histograms show the distribution of scores for each metric across all evaluated segments. The scatter plots illustrate the correlation between automated metric scores and human judgments according to the "group-by-item" segment-level accuracy with tie calibration. Exact correlation values for each language pair and their respective averages are provided above each plot. The background heatmap represents the mean correlation across all language pairs (note: EN–IT is excluded for COMET22 due to a lack of available references).}
    \label{fig:kendall_metrics}
\end{figure*}

\begin{figure*}[ht]
    \begin{center}
        \includegraphics[scale=0.21, trim={0 0cm 0 1.5cm}, clip]{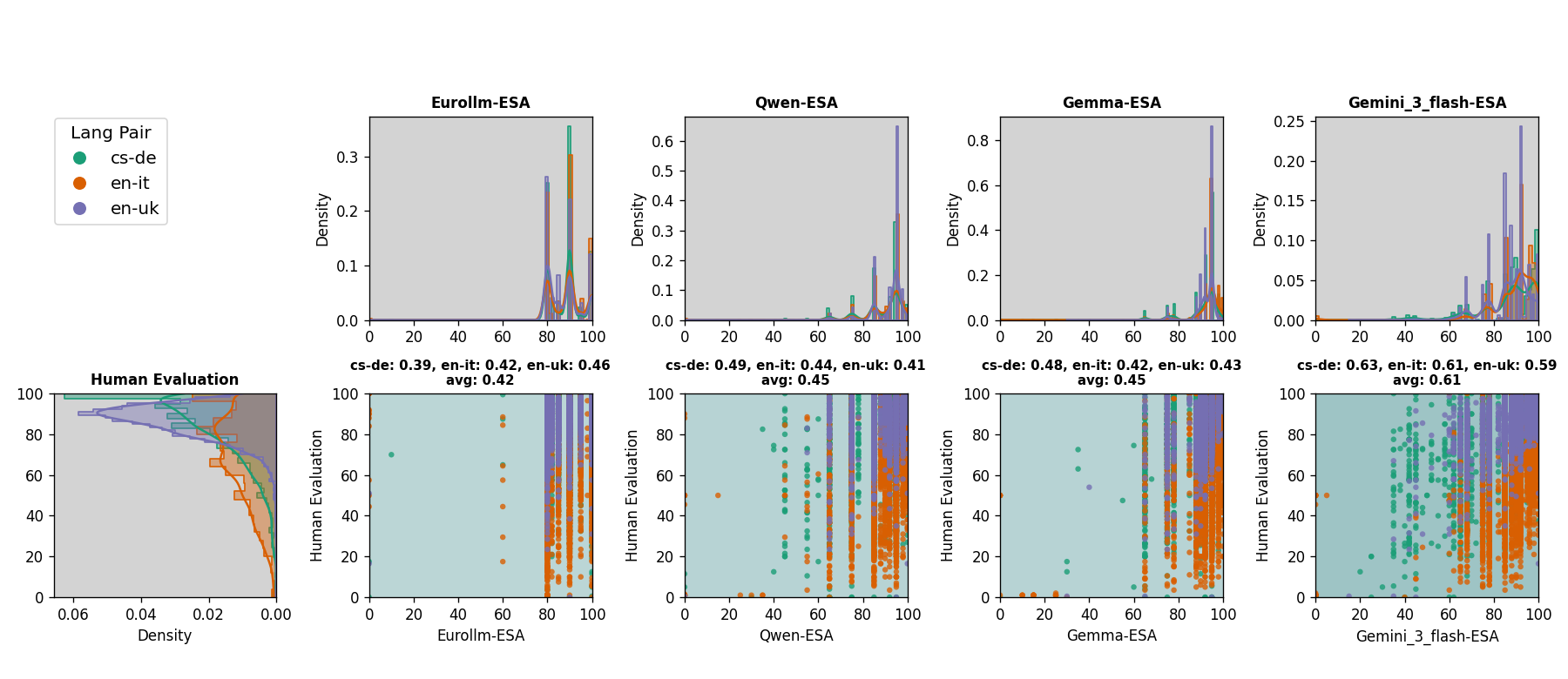}
        \caption{Segment-level performance of the proposed LLM-based QE metrics. The histograms display the distribution of quality scores generated by each model, while the scatter plots demonstrate the correlation with human-assigned ratings using the "group-by-item" segment-level accuracy with tie calibration. Pairwise correlation values for all language pairs and their overall averages are indicated above each plot. The background heatmap represents the mean correlation across all language pairs, with all proposed models providing scores on a scale of 0 to 100.}
        \label{fig:kendall_qe}
    \end{center}
\end{figure*}

\subsection{Correlation with Human ESA}

\begin{figure}[h]
  \includegraphics[width=\linewidth, trim={0 0.8cm 0 2.5cm}, clip]{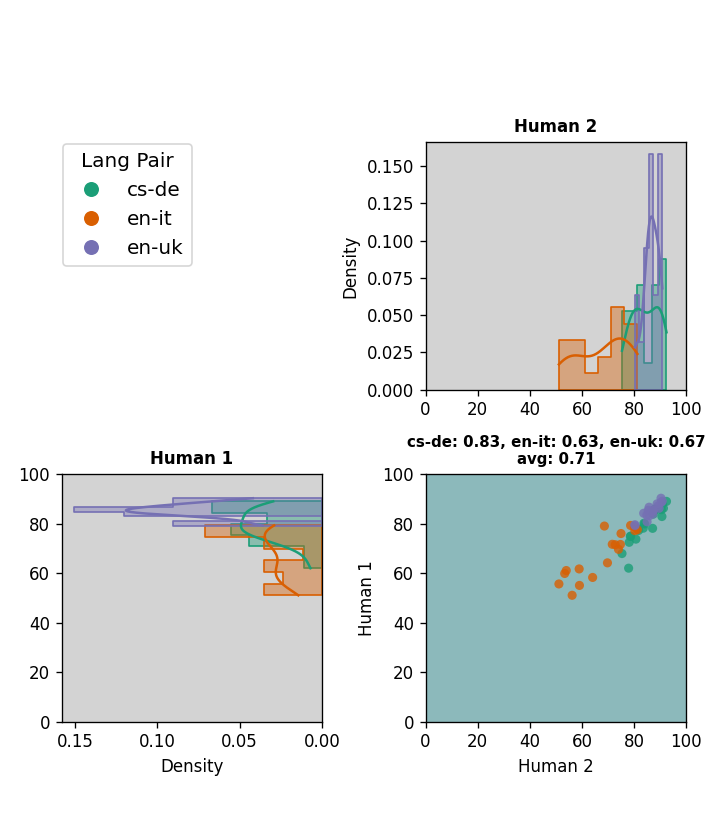}
  \caption{Inter-annotator agreement between independent human evaluators at the system level. The histograms (top and left) display the distribution of average quality scores assigned by each human annotator across all evaluated systems. The scatter plot (bottom-right) illustrates the correlation between the two sets of human scores. Soft Pairwise Accuracy values for each language pair and their overall average are provided above the plot.}
  \label{fig:human-sys}
\end{figure}

\begin{figure}[h]
  \includegraphics[width=\linewidth, trim={0 0.8cm 0 3.2cm}, clip]{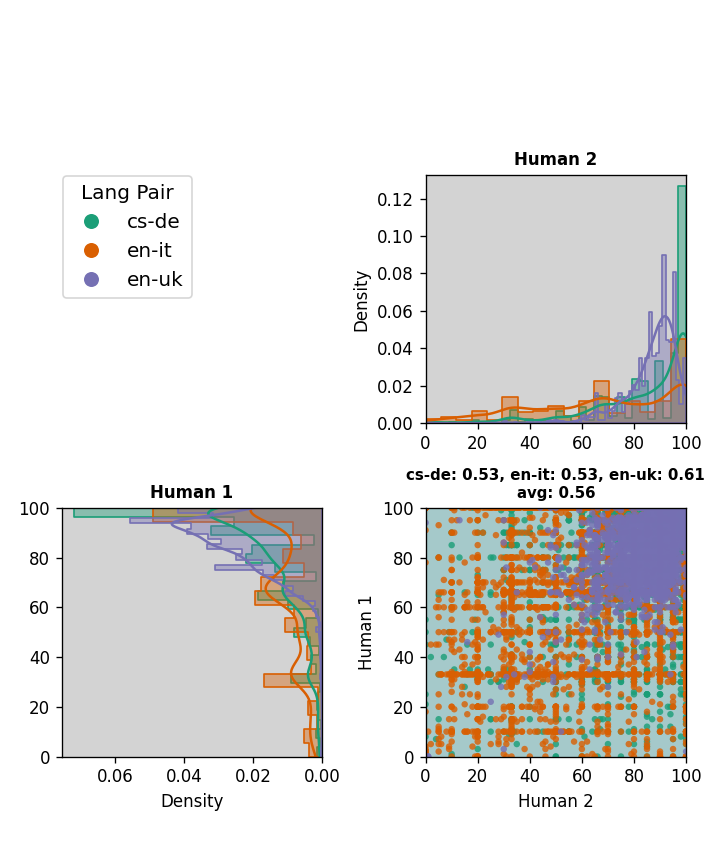}
  \caption{Inter-annotator agreement between independent human evaluators at the segment level. The histograms (top and left) show the distribution of quality scores assigned to individual segments by each human annotator. The scatter plot (bottom-right) demonstrates the correlation between these segment-level judgments. The "group-by-item" segment-level accuracy with tie calibration values for all language pairs and their overall average are provided above the plot.}
  \label{fig:human-seg}
\end{figure}

As illustrated in Figure \ref{fig:pearson_metrics} and Figure \ref{fig:pearson_qe}, the evaluated open-source models demonstrate highly competitive performance at the system level. Specifically, Qwen3 and Gemma achieve strong average SPA scores of 0.83 and 0.82, respectively. While these results are still lower than the proprietary Gemini-2.5-Pro (0.87) and GEMBA-MQM V2 (0.84), they substantially outperform traditional regression metrics such as COMET22 (0.73) and its reference-free counterpart COMETKiwi22 (0.60). Notably, these open-source models also surpass MetricX-25-QE (0.71), a SOTA model explicitly fine-tuned for the QE task. Among the open-weight models, EuroLLM exhibits a slightly lower average SPA of 0.72, though it remains competitive with the baselines.

Conversely, as seen in Figure \ref{fig:kendall_metrics} and Figure \ref{fig:kendall_qe}, the performance at the segment level reveals a persistent limitation of current open-source LLM evaluators. All evaluated open-source models exhibit significantly lower correlations compared to both proprietary LLMs and dedicated neural metrics. Only proprietary models achieve segment-level correlations that approach the reliability of traditional neural metrics. This aligns with previous findings that while LLMs excel at aggregating system-level quality, they struggle to consistently and reliably rank individual segments \cite{lavie-etal-2025-findings}. However, LLM-based evaluators can be viewed as complementary to these traditional metrics. While they may show lower ranking correlation, they provide a higher level of interpretability than regression-based models.

Furthermore, our evaluation highlights visible performance disparities across different language pairs. These variations are present for both the open-source and proprietary models. Because the model performance on specific language pairs differs depending on the pre-training data, carefully selecting and validating the appropriate model remains a crucial step for practical deployment.

Finally, an analysis of the score distributions reveals a pattern for generative models: score quantization. As illustrated in the scatter plots in Figure \ref{fig:kendall_qe}, LLM-generated scores frequently exhibit visible gaps, anchoring to specific numerical values. In contrast, traditional neural metrics yield smooth, continuous score distributions.

Additionally we present the results in tabular form in Appendix \ref{appendix:results}.

\subsection{Inter-Annotator Agreement}
To contextualize the performance of the automated metrics, we computed the inter-annotator agreement between human evaluators who provided annotations for the WMT25 dataset. At the system level, presented in Figure \ref{fig:human-sys}, the average SPA between humans is only 0.71. Remarkably, all evaluated LLM-based QE metrics exceed this, demonstrating that automated metrics can provide more consistent system-level rankings than a secondary human annotator. Furthermore, human agreement exhibits substantial variance across language pairs, mirroring the observations made in automated metrics. 

At the segment level, presented in Figure \ref{fig:human-seg}, the average human correlation drops significantly to 0.56. This confirms that segment-level scoring is an inherently subjective task, making it more difficult to model than system-level scoring. While the open-source models currently fall short in this task, Gemini actually achieves higher agreement with the gold standard.

Finally, the human-to-human comparison provides insight into the score quantization observed in the LLM outputs. The scatter plot in Figure \ref{fig:human-seg} reveals that human evaluators also tend to anchor their assessments to specific round numbers, creating a grid-like visual pattern that is particularly pronounced in the English-Italian data. This indicates that the discretized scoring behavior of generative models is not strictly an artificial flaw, but might reflect a human tendency to categorize subjective quality into distinct, quantized tiers.

\subsection{Error Span Detection}

\begin{table*}[t!h]
\centering
\begin{tabular}{|l|ccc|ccc|} \hline
& \multicolumn{3}{c|}{Unfiltered errors} & \multicolumn{3}{c|}{Filtered errors} \\
& Precision & Recall & F1 score & Precision & Recall & F1 score \\ \hline
\multicolumn{7}{|c|}{\cellcolor{gray!20}\textbf{\small Czech → German}} \\ \hline
EuroLLM-9B-Instruct & 6.21 & 9.03 & \cellcolor{teal!15}7.36 & 6.59 & 8.49 & \cellcolor{teal!15}7.42 \\
Qwen3-VL-30B-A3B-Instruct & 4.55 & \textbf{20.39} & \cellcolor{teal!15}7.43 & 6.87 & 18.41 & \cellcolor{teal!20}10.01 \\
Gemma-3-27b-it & 11.12 & 9.86 & \cellcolor{teal!21}10.45 & 11.24 & 9.62 & \cellcolor{teal!21}10.37 \\
Gemini-3-Flash & \textbf{16.53} & 20.23 & \cellcolor{teal!36}\textbf{18.19} & \textbf{16.66} & \textbf{19.97} & \cellcolor{teal!36}\textbf{18.17} \\ \hline
\multicolumn{7}{|c|}{\cellcolor{gray!20}\textbf{\small English → Italian}} \\ \hline
EuroLLM-9B-Instruct & 4.35 & 7.43 & \cellcolor{teal!11}5.49 & 4.44 & 7.16 & \cellcolor{teal!11}5.48 \\
Qwen3-VL-30B-A3B-Instruct & 3.03 & \textbf{25.75} & \cellcolor{teal!11}5.42 & 4.32 & \textbf{23.42} & \cellcolor{teal!15}7.30 \\
Gemma-3-27b-it & 9.78 & 11.69 & \cellcolor{teal!22}10.65 & 9.82 & 11.47 & \cellcolor{teal!21}10.58 \\
Gemini-3-Flash & \textbf{10.13} & 22.10 & \cellcolor{teal!28}\textbf{13.89} & \textbf{10.20} & 21.96 & \cellcolor{teal!28}\textbf{13.93} \\ \hline
\multicolumn{7}{|c|}{\cellcolor{gray!20}\textbf{\small English → Ukrainian}} \\ \hline
EuroLLM-9B-Instruct & 2.28 & 8.89 & \cellcolor{teal!7}3.63 & 2.30 & 8.41 & \cellcolor{teal!7}3.61 \\
Qwen3-VL-30B-A3B-Instruct & 1.51 & 23.43 & \cellcolor{teal!6}2.84 & 1.84 & 20.65 & \cellcolor{teal!7}3.39 \\
Gemma-3-27b-it & 3.77 & 11.37 & \cellcolor{teal!11}5.67 & 3.76 & 11.11 & \cellcolor{teal!11}5.61 \\
Gemini-3-Flash & \textbf{4.57} & \textbf{24.72} & \cellcolor{teal!15}\textbf{7.72} & \textbf{4.58} & \textbf{24.27} & \cellcolor{teal!15}\textbf{7.71} \\ \hline
\end{tabular}
\caption{Span-level error detection performance across three language pairs, evaluated against official human annotations. The table shows precision, recall, and F1 scores for four QE models, considering both unfiltered and filtered error sets. The highest score for each metric within a language pair is indicated in bold, and a heatmap is applied to the F1 score column.}
\label{tab:accuracy_humeval}
\vspace{0.5cm}
\centering
\begin{tabular}{|l|ccc|ccc|} \hline
& \multicolumn{3}{c|}{Unfiltered errors} & \multicolumn{3}{c|}{Filtered errors} \\
& Precision & Recall & F1 score & Precision & Recall & F1 score \\ \hline
\multicolumn{7}{|c|}{\cellcolor{gray!20}\textbf{\small Czech → German}} \\ \hline
EuroLLM-9B-Instruct & 9.58 & 11.38 & \cellcolor{teal!21}10.40 & 10.16 & 10.92 & \cellcolor{teal!21}10.52 \\
Qwen3-VL-30B-A3B-Instruct & 8.11 & \textbf{29.72} & \cellcolor{teal!25}12.74 & 12.26 & \textbf{27.41} & \cellcolor{teal!34}16.95 \\
Gemma-3-27b-it & \textbf{24.82} & 17.98 & \cellcolor{teal!42}\textbf{20.86} & \textbf{24.83} & 17.73 & \cellcolor{teal!41}\textbf{20.69} \\ \hline
\multicolumn{7}{|c|}{\cellcolor{gray!20}\textbf{\small English → Italian}} \\ \hline
EuroLLM-9B-Instruct & 12.30 & 9.64 & \cellcolor{teal!22}10.81 & 12.54 & 9.40 & \cellcolor{teal!22}10.75 \\
Qwen3-VL-30B-A3B-Instruct & 8.38 & \textbf{32.69} & \cellcolor{teal!27}13.34 & 11.97 & \textbf{30.11} & \cellcolor{teal!34}17.13 \\
Gemma-3-27b-it & \textbf{27.04} & 14.81 & \cellcolor{teal!38}\textbf{19.14} & \textbf{27.10} & 14.71 & \cellcolor{teal!38}\textbf{19.07} \\ \hline
\multicolumn{7}{|c|}{\cellcolor{gray!20}\textbf{\small English → Ukrainian}} \\ \hline
EuroLLM-9B-Instruct & 12.87 & 9.30 & \cellcolor{teal!22}10.80 & 13.17 & 9.10 & \cellcolor{teal!22}10.76 \\
Qwen3-VL-30B-A3B-Instruct & 10.11 & \textbf{28.98} & \cellcolor{teal!30}14.99 & 12.81 & \textbf{27.07} & \cellcolor{teal!35}17.39 \\
Gemma-3-27b-it & \textbf{27.60} & 15.38 & \cellcolor{teal!40}\textbf{19.75} & \textbf{27.43} & 15.30 & \cellcolor{teal!39}\textbf{19.64} \\ \hline
\end{tabular}
\caption{Precision, recall, and F1 scores for span-level error detection, using Gemini-3-Flash predictions as the gold standard. This evaluation measures the alignment of open-source QE models with the proprietary \texttt{Gemini-3-Flash} model across three language pairs. Performance is reported for both unfiltered and filtered error sets, with the highest values per metric bolded and F1 scores visualized via a heatmap.
}
\label{tab:accuracy_gemini}
\end{table*}

As shown in Table \ref{tab:accuracy_humeval}, the absolute F1 scores for error span detection remain consistently low across all models, reaching a maximum of approximately 18\% for the proprietary Gemini-3-Flash and roughly 10\% for the best open-source model, Gemma-3-27b-it. However, evaluating automated error span detection is complicated by inherently low human inter-annotator agreement. For the language pairs examined in our study, the micro-F1 agreement between independent human annotators ranges from only 30\% to 37\% \cite{lavie-etal-2025-findings}, which establishes a low empirical ceiling for the task.

Furthermore, by applying simple error filtering, we can effectively limit the number of annotated errors without any significant decrease in the overall F1 score. This filtering process makes the annotations more practically useful by curbing over-annotation. For instance, Qwen3 exhibits highest recall prior to filtering (e.g., 25.75 on English-Italian) but low precision (3.03), suggesting a strong tendency to over-annotate or hallucinate spans, which is confirmed by the amount of filtered errors presented in Table \ref{tab:errors}. After filtering, Qwen3 improves its precision without sacrificing much recall, increasing its F1 score. Overall, Gemma-3-27b-it proves to be the most capable open-source model for span detection, achieving the highest F1 scores, though this performance may simply be due to the model having the largest number of active parameters.

In addition to evaluating the span-detection metrics against human-annotated error spans, we also measured performance against the error spans generated by Gemini-3-Flash to determine if the open-source models exhibit higher alignment with a state-of-the-art proprietary LLM. As shown in Table \ref{tab:accuracy_gemini}, the open-source models reach a notably higher agreement with Gemini than with human annotators, demonstrating significant increases in both precision and recall across all tested language pairs. Although this model-to-model agreement is still lower than the human inter-annotator agreement, it demonstrates that open-source models may be capable of approximating the proprietary LLMs.

\section{Conclusion}

In this work, we demonstrated the viability of using smaller, open-source LLMs for interpretable translation QE. By utilizing a streamlined prompting strategy, our approach efficiently generates quality scores, MQM-style error spans with suggested corrections, and full segment post-editions in a single pass. This unified approach reduces computational overhead and provides immediate, actionable feedback that is highly valuable for professional translation workflows.

Our evaluation revealed that open-source models are highly effective for system-level ranking. They consistently outperform traditional neural regression metrics, fine-tuned QE models, and even the human inter-annotator agreement. However, segment-level and span-level QE remain a challenge, both for small open-source and larger proprietary LLMs. Our analysis demonstrated that these tasks are highly subjective, resulting in low inter-annotator agreement. To address the tendency of LLMs to over-annotate, we showed that a simple heuristic filtering pipeline can prove useful.

\section{Future Work}

Although our framework generates segment post-editions alongside quality scores and error annotations, we leave evaluating these corrections for future work. It would be beneficial to investigate how explicit error detection affects the quality of these post-editions and how they could be used to refine our filtering pipeline by verifying error spans against the model's suggested corrections to further reduce false positives.

Furthermore, the relationship between the generated error annotations and the final QE scores requires further analysis. Subsequent research could examine how the number and severity of identified errors impact the assigned score and its correlation with human judgment.

\section{Limitations}

The open-source models evaluated in this study are general-purpose multilingual LLMs operating in a few-shot setting. Because they were not explicitly fine-tuned for the QE task, their performance relies entirely on their pre-trained instruction-following capabilities. Task-specific fine-tuning on MQM or ESA datasets could potentially yield significant performance improvements and better alignment with human annotation logic.

Furthermore, to accommodate the VRAM constraints of consumer-grade GPUs, our approach evaluates translation segments in isolation. This omission of document-level context prevents the models from detecting broader discourse-level errors, such as cross-sentence coreference failures. On top of that, our evaluation framework currently does not incorporate external terminological specifications, cultural context, or details regarding the intended target audience. In professional localization workflows, these contextual layers are often critical for determining the true severity of an error or the overall appropriateness of a translation.

Additionally, the character-level F1 metric used for span-level error detection evaluation is strictly based on span overlap and severity weighting. It does not assess the accuracy of the generated error typology (e.g., distinguishing a grammatical error from a mistranslation) or the validity of the suggested corrections. Consequently, the current evaluation paradigm may underestimate the practical utility of the model's output by rewarding only the span of the detected error, while ignoring the usefulness of the provided feedback.

\bibliography{eamt26}

@inproceedings{junczys-dowmunt-2025-gemba,
    title = "{GEMBA} V2: Ten Judgments Are Better Than One",
    author = "Junczys-Dowmunt, Marcin",
    editor = "Haddow, Barry  and
      Kocmi, Tom  and
      Koehn, Philipp  and
      Monz, Christof",
    booktitle = "Proceedings of the Tenth Conference on Machine Translation",
    month = nov,
    year = "2025",
    address = "Suzhou, China",
    publisher = "Association for Computational Linguistics",
    url = "https://aclanthology.org/2025.wmt-1.67/",
    doi = "10.18653/v1/2025.wmt-1.67",
    pages = "926--933",
    ISBN = "979-8-89176-341-8"
}

@misc{gemini3flash2025,
  title = {Gemini 3 {Flash} Model Card},
  author = {{Google DeepMind}},
  year = {2025},
  month = dec,
  howpublished = {\url{https://storage.googleapis.com/deepmind-media/Model-Cards/Gemini-3-Flash-Model-Card.pdf}},
  note = {Accessed: 2026-03-13}
}

@inproceedings{rei-etal-2020-comet,
    title = "{COMET}: A Neural Framework for {MT} Evaluation",
    author = "Rei, Ricardo  and
      Stewart, Craig  and
      Farinha, Ana C  and
      Lavie, Alon",
    editor = "Webber, Bonnie  and
      Cohn, Trevor  and
      He, Yulan  and
      Liu, Yang",
    booktitle = "Proceedings of the 2020 Conference on Empirical Methods in Natural Language Processing (EMNLP)",
    month = nov,
    year = "2020",
    address = "Online",
    publisher = "Association for Computational Linguistics",
    url = "https://aclanthology.org/2020.emnlp-main.213/",
    doi = "10.18653/v1/2020.emnlp-main.213",
    pages = "2685--2702"
}

@inproceedings{kocmi-federmann-2023-gemba,
    title = "{GEMBA}-{MQM}: Detecting Translation Quality Error Spans with {GPT}-4",
    author = "Kocmi, Tom  and
      Federmann, Christian",
    editor = "Koehn, Philipp  and
      Haddow, Barry  and
      Kocmi, Tom  and
      Monz, Christof",
    booktitle = "Proceedings of the Eighth Conference on Machine Translation",
    month = dec,
    year = "2023",
    address = "Singapore",
    publisher = "Association for Computational Linguistics",
    url = "https://aclanthology.org/2023.wmt-1.64/",
    doi = "10.18653/v1/2023.wmt-1.64",
    pages = "768--775"
}

@inproceedings{fernandes-etal-2023-devil,
    title = "The Devil Is in the Errors: Leveraging Large Language Models for Fine-grained Machine Translation Evaluation",
    author = "Fernandes, Patrick  and
      Deutsch, Daniel  and
      Finkelstein, Mara  and
      Riley, Parker  and
      Martins, Andr{\'e}  and
      Neubig, Graham  and
      Garg, Ankush  and
      Clark, Jonathan  and
      Freitag, Markus  and
      Firat, Orhan",
    editor = "Koehn, Philipp  and
      Haddow, Barry  and
      Kocmi, Tom  and
      Monz, Christof",
    booktitle = "Proceedings of the Eighth Conference on Machine Translation",
    month = dec,
    year = "2023",
    address = "Singapore",
    publisher = "Association for Computational Linguistics",
    url = "https://aclanthology.org/2023.wmt-1.100/",
    doi = "10.18653/v1/2023.wmt-1.100",
    pages = "1066--1083"
}

@inproceedings{burchardt-2013-multidimensional,
    title = "Multidimensional quality metrics: a flexible system for assessing translation quality",
    author = "Lommel, Arle Richard  and
      Burchardt, Aljoscha  and
      Uszkoreit, Hans",
    booktitle = "Proceedings of Translating and the Computer 35",
    month = nov # " 28-29",
    year = "2013",
    address = "London, UK",
    publisher = "Aslib",
    url = "https://aclanthology.org/2013.tc-1.6/"
}

@inproceedings{kocmi-etal-2024-error,
    title = "Error Span Annotation: A Balanced Approach for Human Evaluation of Machine Translation",
    author = "Kocmi, Tom  and
      Zouhar, Vil{\'e}m  and
      Avramidis, Eleftherios  and
      Grundkiewicz, Roman  and
      Karpinska, Marzena  and
      Popovi{\'c}, Maja  and
      Sachan, Mrinmaya  and
      Shmatova, Mariya",
    editor = "Haddow, Barry  and
      Kocmi, Tom  and
      Koehn, Philipp  and
      Monz, Christof",
    booktitle = "Proceedings of the Ninth Conference on Machine Translation",
    month = nov,
    year = "2024",
    address = "Miami, Florida, USA",
    publisher = "Association for Computational Linguistics",
    url = "https://aclanthology.org/2024.wmt-1.131/",
    doi = "10.18653/v1/2024.wmt-1.131",
    pages = "1440--1453"
}

@inproceedings{rei-etal-2022-cometkiwi,
    title = "{C}omet{K}iwi: {IST}-Unbabel 2022 Submission for the Quality Estimation Shared Task",
    author = "Rei, Ricardo  and
      Treviso, Marcos  and
      Guerreiro, Nuno M.  and
      Zerva, Chrysoula  and
      Farinha, Ana C  and
      Maroti, Christine  and
      C. de Souza, Jos{\'e} G.  and
      Glushkova, Taisiya  and
      Alves, Duarte  and
      Coheur, Luisa  and
      Lavie, Alon  and
      Martins, Andr{\'e} F. T.",
    editor = {Koehn, Philipp  and
      Barrault, Lo{\"i}c  and
      Bojar, Ond{\v{r}}ej  and
      Bougares, Fethi  and
      Chatterjee, Rajen  and
      Costa-juss{\`a}, Marta R.  and
      Federmann, Christian  and
      Fishel, Mark  and
      Fraser, Alexander  and
      Freitag, Markus  and
      Graham, Yvette  and
      Grundkiewicz, Roman  and
      Guzman, Paco  and
      Haddow, Barry  and
      Huck, Matthias  and
      Jimeno Yepes, Antonio  and
      Kocmi, Tom  and
      Martins, Andr{\'e}  and
      Morishita, Makoto  and
      Monz, Christof  and
      Nagata, Masaaki  and
      Nakazawa, Toshiaki  and
      Negri, Matteo  and
      N{\'e}v{\'e}ol, Aur{\'e}lie  and
      Neves, Mariana  and
      Popel, Martin  and
      Turchi, Marco  and
      Zampieri, Marcos},
    booktitle = "Proceedings of the Seventh Conference on Machine Translation (WMT)",
    month = dec,
    year = "2022",
    address = "Abu Dhabi, United Arab Emirates (Hybrid)",
    publisher = "Association for Computational Linguistics",
    url = "https://aclanthology.org/2022.wmt-1.60/",
    doi = "10.18653/v1/2022.wmt-1.60",
    pages = "634--645"
}

@article{guerreiro-etal-2024-xcomet,
    title = "x{COMET}: Transparent Machine Translation Evaluation through Fine-grained Error Detection",
    author = "Guerreiro, Nuno M.  and
      Rei, Ricardo  and
      Stigt, Daan van  and
      Coheur, Luisa  and
      Colombo, Pierre  and
      Martins, Andr{\'e} F. T.",
    journal = "Transactions of the Association for Computational Linguistics",
    volume = "12",
    year = "2024",
    address = "Cambridge, MA",
    publisher = "MIT Press",
    url = "https://aclanthology.org/2024.tacl-1.54/",
    doi = "10.1162/tacl_a_00683",
    pages = "979--995"
}

@inproceedings{lu-etal-2024-error,
    title = "Error Analysis Prompting Enables Human-Like Translation Evaluation in Large Language Models",
    author = "Lu, Qingyu  and
      Qiu, Baopu  and
      Ding, Liang  and
      Zhang, Kanjian  and
      Kocmi, Tom  and
      Tao, Dacheng",
    editor = "Ku, Lun-Wei  and
      Martins, Andre  and
      Srikumar, Vivek",
    booktitle = "Findings of the Association for Computational Linguistics: ACL 2024",
    month = aug,
    year = "2024",
    address = "Bangkok, Thailand",
    publisher = "Association for Computational Linguistics",
    url = "https://aclanthology.org/2024.findings-acl.520/",
    doi = "10.18653/v1/2024.findings-acl.520",
    pages = "8801--8816"
}

@inproceedings{lu-etal-2025-mqm,
    title = "{MQM}-{APE}: Toward High-Quality Error Annotation Predictors with Automatic Post-Editing in {LLM} Translation Evaluators",
    author = "Lu, Qingyu  and
      Ding, Liang  and
      Zhang, Kanjian  and
      Zhang, Jinxia  and
      Tao, Dacheng",
    editor = "Rambow, Owen  and
      Wanner, Leo  and
      Apidianaki, Marianna  and
      Al-Khalifa, Hend  and
      Eugenio, Barbara Di  and
      Schockaert, Steven",
    booktitle = "Proceedings of the 31st International Conference on Computational Linguistics",
    month = jan,
    year = "2025",
    address = "Abu Dhabi, UAE",
    publisher = "Association for Computational Linguistics",
    url = "https://aclanthology.org/2025.coling-main.374/",
    pages = "5570--5587"
}

@misc{grattafiori2024llama3herdmodels,
      title={The Llama 3 Herd of Models}, 
      author={Aaron Grattafiori and Abhimanyu Dubey and Abhinav Jauhri and Abhinav Pandey and Abhishek Kadian and Ahmad Al-Dahle and Aiesha Letman and Akhil Mathur and Alan Schelten and Alex Vaughan and Amy Yang and Angela Fan and Anirudh Goyal and Anthony Hartshorn and Aobo Yang and Archi Mitra and Archie Sravankumar and Artem Korenev and Arthur Hinsvark and Arun Rao and Aston Zhang and Aurelien Rodriguez and Austen Gregerson and Ava Spataru and Baptiste Roziere and Bethany Biron and Binh Tang and Bobbie Chern and Charlotte Caucheteux and Chaya Nayak and Chloe Bi and Chris Marra and Chris McConnell and Christian Keller and Christophe Touret and Chunyang Wu and Corinne Wong and Cristian Canton Ferrer and Cyrus Nikolaidis and Damien Allonsius and Daniel Song and Danielle Pintz and Danny Livshits and Danny Wyatt and David Esiobu and Dhruv Choudhary and Dhruv Mahajan and Diego Garcia-Olano and Diego Perino and Dieuwke Hupkes and Egor Lakomkin and Ehab AlBadawy and Elina Lobanova and Emily Dinan and Eric Michael Smith and Filip Radenovic and Francisco Guzmán and Frank Zhang and Gabriel Synnaeve and Gabrielle Lee and Georgia Lewis Anderson and Govind Thattai and Graeme Nail and Gregoire Mialon and Guan Pang and Guillem Cucurell and Hailey Nguyen and Hannah Korevaar and Hu Xu and Hugo Touvron and Iliyan Zarov and Imanol Arrieta Ibarra and Isabel Kloumann and Ishan Misra and Ivan Evtimov and Jack Zhang and Jade Copet and Jaewon Lee and Jan Geffert and Jana Vranes and Jason Park and Jay Mahadeokar and Jeet Shah and Jelmer van der Linde and Jennifer Billock and Jenny Hong and Jenya Lee and Jeremy Fu and Jianfeng Chi and Jianyu Huang and Jiawen Liu and Jie Wang and Jiecao Yu and Joanna Bitton and Joe Spisak and Jongsoo Park and Joseph Rocca and Joshua Johnstun and Joshua Saxe and Junteng Jia and Kalyan Vasuden Alwala and Karthik Prasad and Kartikeya Upasani and Kate Plawiak and Ke Li and Kenneth Heafield and Kevin Stone and Khalid El-Arini and Krithika Iyer and Kshitiz Malik and Kuenley Chiu and Kunal Bhalla and Kushal Lakhotia and Lauren Rantala-Yeary and Laurens van der Maaten and Lawrence Chen and Liang Tan and Liz Jenkins and Louis Martin and Lovish Madaan and Lubo Malo and Lukas Blecher and Lukas Landzaat and Luke de Oliveira and Madeline Muzzi and Mahesh Pasupuleti and Mannat Singh and Manohar Paluri and Marcin Kardas and Maria Tsimpoukelli and Mathew Oldham and Mathieu Rita and Maya Pavlova and Melanie Kambadur and Mike Lewis and Min Si and Mitesh Kumar Singh and Mona Hassan and Naman Goyal and Narjes Torabi and Nikolay Bashlykov and Nikolay Bogoychev and Niladri Chatterji and Ning Zhang and Olivier Duchenne and Onur Çelebi and Patrick Alrassy and Pengchuan Zhang and Pengwei Li and Petar Vasic and Peter Weng and Prajjwal Bhargava and Pratik Dubal and Praveen Krishnan and Punit Singh Koura and Puxin Xu and Qing He and Qingxiao Dong and Ragavan Srinivasan and Raj Ganapathy and Ramon Calderer and Ricardo Silveira Cabral and Robert Stojnic and Roberta Raileanu and Rohan Maheswari and Rohit Girdhar and Rohit Patel and Romain Sauvestre and Ronnie Polidoro and Roshan Sumbaly and Ross Taylor and Ruan Silva and Rui Hou and Rui Wang and Saghar Hosseini and Sahana Chennabasappa and Sanjay Singh and Sean Bell and Seohyun Sonia Kim and Sergey Edunov and Shaoliang Nie and Sharan Narang and Sharath Raparthy and Sheng Shen and Shengye Wan and Shruti Bhosale and Shun Zhang and Simon Vandenhende and Soumya Batra and Spencer Whitman and Sten Sootla and Stephane Collot and Suchin Gururangan and Sydney Borodinsky and Tamar Herman and Tara Fowler and Tarek Sheasha and Thomas Georgiou and Thomas Scialom and Tobias Speckbacher and Todor Mihaylov and Tong Xiao and Ujjwal Karn and Vedanuj Goswami and Vibhor Gupta and Vignesh Ramanathan and Viktor Kerkez and Vincent Gonguet and Virginie Do and Vish Vogeti and Vítor Albiero and Vladan Petrovic and Weiwei Chu and Wenhan Xiong and Wenyin Fu and Whitney Meers and Xavier Martinet and Xiaodong Wang and Xiaofang Wang and Xiaoqing Ellen Tan and Xide Xia and Xinfeng Xie and Xuchao Jia and Xuewei Wang and Yaelle Goldschlag and Yashesh Gaur and Yasmine Babaei and Yi Wen and Yiwen Song and Yuchen Zhang and Yue Li and Yuning Mao and Zacharie Delpierre Coudert and Zheng Yan and Zhengxing Chen and Zoe Papakipos and Aaditya Singh and Aayushi Srivastava and Abha Jain and Adam Kelsey and Adam Shajnfeld and Adithya Gangidi and Adolfo Victoria and Ahuva Goldstand and Ajay Menon and Ajay Sharma and Alex Boesenberg and Alexei Baevski and Allie Feinstein and Amanda Kallet and Amit Sangani and Amos Teo and Anam Yunus and Andrei Lupu and Andres Alvarado and Andrew Caples and Andrew Gu and Andrew Ho and Andrew Poulton and Andrew Ryan and Ankit Ramchandani and Annie Dong and Annie Franco and Anuj Goyal and Aparajita Saraf and Arkabandhu Chowdhury and Ashley Gabriel and Ashwin Bharambe and Assaf Eisenman and Azadeh Yazdan and Beau James and Ben Maurer and Benjamin Leonhardi and Bernie Huang and Beth Loyd and Beto De Paola and Bhargavi Paranjape and Bing Liu and Bo Wu and Boyu Ni and Braden Hancock and Bram Wasti and Brandon Spence and Brani Stojkovic and Brian Gamido and Britt Montalvo and Carl Parker and Carly Burton and Catalina Mejia and Ce Liu and Changhan Wang and Changkyu Kim and Chao Zhou and Chester Hu and Ching-Hsiang Chu and Chris Cai and Chris Tindal and Christoph Feichtenhofer and Cynthia Gao and Damon Civin and Dana Beaty and Daniel Kreymer and Daniel Li and David Adkins and David Xu and Davide Testuggine and Delia David and Devi Parikh and Diana Liskovich and Didem Foss and Dingkang Wang and Duc Le and Dustin Holland and Edward Dowling and Eissa Jamil and Elaine Montgomery and Eleonora Presani and Emily Hahn and Emily Wood and Eric-Tuan Le and Erik Brinkman and Esteban Arcaute and Evan Dunbar and Evan Smothers and Fei Sun and Felix Kreuk and Feng Tian and Filippos Kokkinos and Firat Ozgenel and Francesco Caggioni and Frank Kanayet and Frank Seide and Gabriela Medina Florez and Gabriella Schwarz and Gada Badeer and Georgia Swee and Gil Halpern and Grant Herman and Grigory Sizov and Guangyi and Zhang and Guna Lakshminarayanan and Hakan Inan and Hamid Shojanazeri and Han Zou and Hannah Wang and Hanwen Zha and Haroun Habeeb and Harrison Rudolph and Helen Suk and Henry Aspegren and Hunter Goldman and Hongyuan Zhan and Ibrahim Damlaj and Igor Molybog and Igor Tufanov and Ilias Leontiadis and Irina-Elena Veliche and Itai Gat and Jake Weissman and James Geboski and James Kohli and Janice Lam and Japhet Asher and Jean-Baptiste Gaya and Jeff Marcus and Jeff Tang and Jennifer Chan and Jenny Zhen and Jeremy Reizenstein and Jeremy Teboul and Jessica Zhong and Jian Jin and Jingyi Yang and Joe Cummings and Jon Carvill and Jon Shepard and Jonathan McPhie and Jonathan Torres and Josh Ginsburg and Junjie Wang and Kai Wu and Kam Hou U and Karan Saxena and Kartikay Khandelwal and Katayoun Zand and Kathy Matosich and Kaushik Veeraraghavan and Kelly Michelena and Keqian Li and Kiran Jagadeesh and Kun Huang and Kunal Chawla and Kyle Huang and Lailin Chen and Lakshya Garg and Lavender A and Leandro Silva and Lee Bell and Lei Zhang and Liangpeng Guo and Licheng Yu and Liron Moshkovich and Luca Wehrstedt and Madian Khabsa and Manav Avalani and Manish Bhatt and Martynas Mankus and Matan Hasson and Matthew Lennie and Matthias Reso and Maxim Groshev and Maxim Naumov and Maya Lathi and Meghan Keneally and Miao Liu and Michael L. Seltzer and Michal Valko and Michelle Restrepo and Mihir Patel and Mik Vyatskov and Mikayel Samvelyan and Mike Clark and Mike Macey and Mike Wang and Miquel Jubert Hermoso and Mo Metanat and Mohammad Rastegari and Munish Bansal and Nandhini Santhanam and Natascha Parks and Natasha White and Navyata Bawa and Nayan Singhal and Nick Egebo and Nicolas Usunier and Nikhil Mehta and Nikolay Pavlovich Laptev and Ning Dong and Norman Cheng and Oleg Chernoguz and Olivia Hart and Omkar Salpekar and Ozlem Kalinli and Parkin Kent and Parth Parekh and Paul Saab and Pavan Balaji and Pedro Rittner and Philip Bontrager and Pierre Roux and Piotr Dollar and Polina Zvyagina and Prashant Ratanchandani and Pritish Yuvraj and Qian Liang and Rachad Alao and Rachel Rodriguez and Rafi Ayub and Raghotham Murthy and Raghu Nayani and Rahul Mitra and Rangaprabhu Parthasarathy and Raymond Li and Rebekkah Hogan and Robin Battey and Rocky Wang and Russ Howes and Ruty Rinott and Sachin Mehta and Sachin Siby and Sai Jayesh Bondu and Samyak Datta and Sara Chugh and Sara Hunt and Sargun Dhillon and Sasha Sidorov and Satadru Pan and Saurabh Mahajan and Saurabh Verma and Seiji Yamamoto and Sharadh Ramaswamy and Shaun Lindsay and Shaun Lindsay and Sheng Feng and Shenghao Lin and Shengxin Cindy Zha and Shishir Patil and Shiva Shankar and Shuqiang Zhang and Shuqiang Zhang and Sinong Wang and Sneha Agarwal and Soji Sajuyigbe and Soumith Chintala and Stephanie Max and Stephen Chen and Steve Kehoe and Steve Satterfield and Sudarshan Govindaprasad and Sumit Gupta and Summer Deng and Sungmin Cho and Sunny Virk and Suraj Subramanian and Sy Choudhury and Sydney Goldman and Tal Remez and Tamar Glaser and Tamara Best and Thilo Koehler and Thomas Robinson and Tianhe Li and Tianjun Zhang and Tim Matthews and Timothy Chou and Tzook Shaked and Varun Vontimitta and Victoria Ajayi and Victoria Montanez and Vijai Mohan and Vinay Satish Kumar and Vishal Mangla and Vlad Ionescu and Vlad Poenaru and Vlad Tiberiu Mihailescu and Vladimir Ivanov and Wei Li and Wenchen Wang and Wenwen Jiang and Wes Bouaziz and Will Constable and Xiaocheng Tang and Xiaojian Wu and Xiaolan Wang and Xilun Wu and Xinbo Gao and Yaniv Kleinman and Yanjun Chen and Ye Hu and Ye Jia and Ye Qi and Yenda Li and Yilin Zhang and Ying Zhang and Yossi Adi and Youngjin Nam and Yu and Wang and Yu Zhao and Yuchen Hao and Yundi Qian and Yunlu Li and Yuzi He and Zach Rait and Zachary DeVito and Zef Rosnbrick and Zhaoduo Wen and Zhenyu Yang and Zhiwei Zhao and Zhiyu Ma},
      year={2024},
      eprint={2407.21783},
      archivePrefix={arXiv},
      primaryClass={cs.AI},
      url={https://arxiv.org/abs/2407.21783}, 
}

@misc{jiang2024mixtralexperts,
      title={Mixtral of Experts}, 
      author={Albert Q. Jiang and Alexandre Sablayrolles and Antoine Roux and Arthur Mensch and Blanche Savary and Chris Bamford and Devendra Singh Chaplot and Diego de las Casas and Emma Bou Hanna and Florian Bressand and Gianna Lengyel and Guillaume Bour and Guillaume Lample and Lélio Renard Lavaud and Lucile Saulnier and Marie-Anne Lachaux and Pierre Stock and Sandeep Subramanian and Sophia Yang and Szymon Antoniak and Teven Le Scao and Théophile Gervet and Thibaut Lavril and Thomas Wang and Timothée Lacroix and William El Sayed},
      year={2024},
      eprint={2401.04088},
      archivePrefix={arXiv},
      primaryClass={cs.LG},
      url={https://arxiv.org/abs/2401.04088}, 
}

@misc{gemmateam2025gemma3technicalreport,
      title={Gemma 3 Technical Report}, 
      author={Gemma Team and Aishwarya Kamath and Johan Ferret and Shreya Pathak and Nino Vieillard and Ramona Merhej and Sarah Perrin and Tatiana Matejovicova and Alexandre Ramé and Morgane Rivière and Louis Rouillard and Thomas Mesnard and Geoffrey Cideron and Jean-bastien Grill and Sabela Ramos and Edouard Yvinec and Michelle Casbon and Etienne Pot and Ivo Penchev and Gaël Liu and Francesco Visin and Kathleen Kenealy and Lucas Beyer and Xiaohai Zhai and Anton Tsitsulin and Robert Busa-Fekete and Alex Feng and Noveen Sachdeva and Benjamin Coleman and Yi Gao and Basil Mustafa and Iain Barr and Emilio Parisotto and David Tian and Matan Eyal and Colin Cherry and Jan-Thorsten Peter and Danila Sinopalnikov and Surya Bhupatiraju and Rishabh Agarwal and Mehran Kazemi and Dan Malkin and Ravin Kumar and David Vilar and Idan Brusilovsky and Jiaming Luo and Andreas Steiner and Abe Friesen and Abhanshu Sharma and Abheesht Sharma and Adi Mayrav Gilady and Adrian Goedeckemeyer and Alaa Saade and Alex Feng and Alexander Kolesnikov and Alexei Bendebury and Alvin Abdagic and Amit Vadi and András György and André Susano Pinto and Anil Das and Ankur Bapna and Antoine Miech and Antoine Yang and Antonia Paterson and Ashish Shenoy and Ayan Chakrabarti and Bilal Piot and Bo Wu and Bobak Shahriari and Bryce Petrini and Charlie Chen and Charline Le Lan and Christopher A. Choquette-Choo and CJ Carey and Cormac Brick and Daniel Deutsch and Danielle Eisenbud and Dee Cattle and Derek Cheng and Dimitris Paparas and Divyashree Shivakumar Sreepathihalli and Doug Reid and Dustin Tran and Dustin Zelle and Eric Noland and Erwin Huizenga and Eugene Kharitonov and Frederick Liu and Gagik Amirkhanyan and Glenn Cameron and Hadi Hashemi and Hanna Klimczak-Plucińska and Harman Singh and Harsh Mehta and Harshal Tushar Lehri and Hussein Hazimeh and Ian Ballantyne and Idan Szpektor and Ivan Nardini and Jean Pouget-Abadie and Jetha Chan and Joe Stanton and John Wieting and Jonathan Lai and Jordi Orbay and Joseph Fernandez and Josh Newlan and Ju-yeong Ji and Jyotinder Singh and Kat Black and Kathy Yu and Kevin Hui and Kiran Vodrahalli and Klaus Greff and Linhai Qiu and Marcella Valentine and Marina Coelho and Marvin Ritter and Matt Hoffman and Matthew Watson and Mayank Chaturvedi and Michael Moynihan and Min Ma and Nabila Babar and Natasha Noy and Nathan Byrd and Nick Roy and Nikola Momchev and Nilay Chauhan and Noveen Sachdeva and Oskar Bunyan and Pankil Botarda and Paul Caron and Paul Kishan Rubenstein and Phil Culliton and Philipp Schmid and Pier Giuseppe Sessa and Pingmei Xu and Piotr Stanczyk and Pouya Tafti and Rakesh Shivanna and Renjie Wu and Renke Pan and Reza Rokni and Rob Willoughby and Rohith Vallu and Ryan Mullins and Sammy Jerome and Sara Smoot and Sertan Girgin and Shariq Iqbal and Shashir Reddy and Shruti Sheth and Siim Põder and Sijal Bhatnagar and Sindhu Raghuram Panyam and Sivan Eiger and Susan Zhang and Tianqi Liu and Trevor Yacovone and Tyler Liechty and Uday Kalra and Utku Evci and Vedant Misra and Vincent Roseberry and Vlad Feinberg and Vlad Kolesnikov and Woohyun Han and Woosuk Kwon and Xi Chen and Yinlam Chow and Yuvein Zhu and Zichuan Wei and Zoltan Egyed and Victor Cotruta and Minh Giang and Phoebe Kirk and Anand Rao and Kat Black and Nabila Babar and Jessica Lo and Erica Moreira and Luiz Gustavo Martins and Omar Sanseviero and Lucas Gonzalez and Zach Gleicher and Tris Warkentin and Vahab Mirrokni and Evan Senter and Eli Collins and Joelle Barral and Zoubin Ghahramani and Raia Hadsell and Yossi Matias and D. Sculley and Slav Petrov and Noah Fiedel and Noam Shazeer and Oriol Vinyals and Jeff Dean and Demis Hassabis and Koray Kavukcuoglu and Clement Farabet and Elena Buchatskaya and Jean-Baptiste Alayrac and Rohan Anil and Dmitry and Lepikhin and Sebastian Borgeaud and Olivier Bachem and Armand Joulin and Alek Andreev and Cassidy Hardin and Robert Dadashi and Léonard Hussenot},
      year={2025},
      eprint={2503.19786},
      archivePrefix={arXiv},
      primaryClass={cs.CL},
      url={https://arxiv.org/abs/2503.19786}, 
}

@misc{martins2025eurollm9btechnicalreport,
      title={EuroLLM-9B: Technical Report}, 
      author={Pedro Henrique Martins and João Alves and Patrick Fernandes and Nuno M. Guerreiro and Ricardo Rei and Amin Farajian and Mateusz Klimaszewski and Duarte M. Alves and José Pombal and Nicolas Boizard and Manuel Faysse and Pierre Colombo and François Yvon and Barry Haddow and José G. C. de Souza and Alexandra Birch and André F. T. Martins},
      year={2025},
      eprint={2506.04079},
      archivePrefix={arXiv},
      primaryClass={cs.CL},
      url={https://arxiv.org/abs/2506.04079}, 
}

@misc{bai2025qwen3vltechnicalreport,
      title={Qwen3-VL Technical Report}, 
      author={Shuai Bai and Yuxuan Cai and Ruizhe Chen and Keqin Chen and Xionghui Chen and Zesen Cheng and Lianghao Deng and Wei Ding and Chang Gao and Chunjiang Ge and Wenbin Ge and Zhifang Guo and Qidong Huang and Jie Huang and Fei Huang and Binyuan Hui and Shutong Jiang and Zhaohai Li and Mingsheng Li and Mei Li and Kaixin Li and Zicheng Lin and Junyang Lin and Xuejing Liu and Jiawei Liu and Chenglong Liu and Yang Liu and Dayiheng Liu and Shixuan Liu and Dunjie Lu and Ruilin Luo and Chenxu Lv and Rui Men and Lingchen Meng and Xuancheng Ren and Xingzhang Ren and Sibo Song and Yuchong Sun and Jun Tang and Jianhong Tu and Jianqiang Wan and Peng Wang and Pengfei Wang and Qiuyue Wang and Yuxuan Wang and Tianbao Xie and Yiheng Xu and Haiyang Xu and Jin Xu and Zhibo Yang and Mingkun Yang and Jianxin Yang and An Yang and Bowen Yu and Fei Zhang and Hang Zhang and Xi Zhang and Bo Zheng and Humen Zhong and Jingren Zhou and Fan Zhou and Jing Zhou and Yuanzhi Zhu and Ke Zhu},
      year={2025},
      eprint={2511.21631},
      archivePrefix={arXiv},
      primaryClass={cs.CV},
      url={https://arxiv.org/abs/2511.21631}, 
}

@inproceedings{lavie-etal-2025-findings,
    title = "Findings of the {WMT}25 Shared Task on Automated Translation Evaluation Systems: Linguistic Diversity is Challenging and References Still Help",
    author = "Lavie, Alon  and
      Hanneman, Greg  and
      Agrawal, Sweta  and
      Kanojia, Diptesh  and
      Lo, Chi-Kiu  and
      Zouhar, Vil{\'e}m  and
      Blain, Frederic  and
      Zerva, Chrysoula  and
      Avramidis, Eleftherios  and
      Deoghare, Sourabh  and
      Sindhujan, Archchana  and
      Wang, Jiayi  and
      Adelani, David Ifeoluwa  and
      Thompson, Brian  and
      Kocmi, Tom  and
      Freitag, Markus  and
      Deutsch, Daniel",
    editor = "Haddow, Barry  and
      Kocmi, Tom  and
      Koehn, Philipp  and
      Monz, Christof",
    booktitle = "Proceedings of the Tenth Conference on Machine Translation",
    month = nov,
    year = "2025",
    address = "Suzhou, China",
    publisher = "Association for Computational Linguistics",
    url = "https://aclanthology.org/2025.wmt-1.24/",
    doi = "10.18653/v1/2025.wmt-1.24",
    pages = "436--483",
    ISBN = "979-8-89176-341-8"
}

@inproceedings{kocmi-federmann-2023-large,
    title = "Large Language Models Are State-of-the-Art Evaluators of Translation Quality",
    author = "Kocmi, Tom  and
      Federmann, Christian",
    editor = "Nurminen, Mary  and
      Brenner, Judith  and
      Koponen, Maarit  and
      Latomaa, Sirkku  and
      Mikhailov, Mikhail  and
      Schierl, Frederike  and
      Ranasinghe, Tharindu  and
      Vanmassenhove, Eva  and
      Vidal, Sergi Alvarez  and
      Aranberri, Nora  and
      Nunziatini, Mara  and
      Escart{\'i}n, Carla Parra  and
      Forcada, Mikel  and
      Popovic, Maja  and
      Scarton, Carolina  and
      Moniz, Helena",
    booktitle = "Proceedings of the 24th Annual Conference of the European Association for Machine Translation",
    month = jun,
    year = "2023",
    address = "Tampere, Finland",
    publisher = "European Association for Machine Translation",
    url = "https://aclanthology.org/2023.eamt-1.19/",
    pages = "193--203"
}

@inproceedings{thompson-etal-2024-improving,
    title = "Improving Statistical Significance in Human Evaluation of Automatic Metrics via Soft Pairwise Accuracy",
    author = "Thompson, Brian  and
      Mathur, Nitika  and
      Deutsch, Daniel  and
      Khayrallah, Huda",
    editor = "Haddow, Barry  and
      Kocmi, Tom  and
      Koehn, Philipp  and
      Monz, Christof",
    booktitle = "Proceedings of the Ninth Conference on Machine Translation",
    month = nov,
    year = "2024",
    address = "Miami, Florida, USA",
    publisher = "Association for Computational Linguistics",
    url = "https://aclanthology.org/2024.wmt-1.118/",
    doi = "10.18653/v1/2024.wmt-1.118",
    pages = "1222--1234"
}

@inproceedings{deutsch-etal-2023-ties,
    title = "Ties Matter: Meta-Evaluating Modern Metrics with Pairwise Accuracy and Tie Calibration",
    author = "Deutsch, Daniel  and
      Foster, George  and
      Freitag, Markus",
    editor = "Bouamor, Houda  and
      Pino, Juan  and
      Bali, Kalika",
    booktitle = "Proceedings of the 2023 Conference on Empirical Methods in Natural Language Processing",
    month = dec,
    year = "2023",
    address = "Singapore",
    publisher = "Association for Computational Linguistics",
    url = "https://aclanthology.org/2023.emnlp-main.798/",
    doi = "10.18653/v1/2023.emnlp-main.798",
    pages = "12914--12929"
}

@inproceedings{juraska-etal-2025-metricx,
    title = "{M}etric{X}-25 and {G}em{S}pan{E}val: {G}oogle {T}ranslate Submissions to the {WMT}25 Evaluation Shared Task",
    author = "Juraska, Juraj  and
      Domhan, Tobias  and
      Finkelstein, Mara  and
      Nakagawa, Tetsuji  and
      Kovacs, Geza  and
      Deutsch, Daniel  and
      Wang, Pidong  and
      Freitag, Markus",
    editor = "Haddow, Barry  and
      Kocmi, Tom  and
      Koehn, Philipp  and
      Monz, Christof",
    booktitle = "Proceedings of the Tenth Conference on Machine Translation",
    month = nov,
    year = "2025",
    address = "Suzhou, China",
    publisher = "Association for Computational Linguistics",
    url = "https://aclanthology.org/2025.wmt-1.70/",
    doi = "10.18653/v1/2025.wmt-1.70",
    pages = "957--968",
    ISBN = "979-8-89176-341-8"
}

@software{Baccianella_JSON_Repair_-_2025,
    author  = "Stefano {Baccianella}",
    title   = "JSON Repair - A python module to repair invalid JSON, commonly used to parse the output of LLMs",
    url     = "https://github.com/mangiucugna/json_repair",
    version = "0.55.2",
    year    = 2025
}
\bibliographystyle{eamt26}

\onecolumn
\clearpage
\newpage
\appendix

\begin{figure*}[h]
\section{Prompts}\label{appendix:prompt}

\begin{lstlisting}[captionpos=b, label=lst:system-prompt, aboveskip=0pt]
You are an AI assistant specialized in {source_language}-to-{target_language} translation quality assurance.

You will receive a pair of paragraphs as a JSON structure. For each input, reply with an extended JSON object that contains the information described below.

First, rate the quality of the translation on a scale from 1 to 100, taking the source text into account. If the translation is fully accurate and consistent with the source sentence, return 100. If the translation is not fully accurate and consistent with the source sentence, return a score between 1 and 100 depending on the quality of the translation and the errors present in the translation. If the translation is in a different language than requested, return 0.
Then post-edit the translation. Correct any errors or unnatural phrasing, ensuring accuracy, fluency, and fidelity to the source text. If the original translation is already correct, leave it unchanged.
Then focus on particular errors in the original translation, that you have corrected in the post-edited translation. Each error is classified as one of three categories: "critical", "major", and "minor". Critical errors inhibit comprehension of the text, change the meaning, provide false information, or make the text impossible to understand. Major errors disrupt the flow, make the text difficult to read or awkward, but the original meaning is still recoverable and understandable. Minor errors are technically incorrect, such as typos or punctuation, but do not affect the meaning, readability, or flow of the translated text. For every of the main three categories, additionally identify error types in the translation and sub-classify them. The types of errors are: "accuracy" ("addition", "mistranslation", "omission", "untranslated text"), "fluency" ("character encoding", "grammar", "inconsistency", "punctuation", "register", "spelling"), "style" ("awkward"), "terminology" ("inappropriate for context", "inconsistent use"), "non-translation", or "other". For every error type, supply suggested correction of the error ("correction") and a very short and concise description of the error ("short_desc"). The correction should be a single suggested word or phrase that is a direct replacement for the error. If there are no errors of a specific main category (critical, major or minor), it is OK to return an empty list for that category. It also OK to not return any errors for any category if everything is fine.

Here is an example of a JSON input for English-German translation:
{
    "source_language": "English",
    "source": "I do apologise about this, we must gain permission from the account holder to discuss an order with another person, I apologise if this was done previously, however, I would not be able to discuss this with yourself without the account holders permission.",
    "translation_language": "German",
    "translation": "Ich entschuldige mich dafür, wir müssen die Erlaubnis einholen, um eine Bestellung mit einer anderen Person zu involvment. Ich entschuldige mich, falls dies zuvor geschehen wäre, aber ohne die Erlaubnis des Kontoinhabers wäre ich nicht in der Lage, dies mit dir involvement."
}

And here is a corresponding JSON output with the score, post-edited translation and error annotations:
{
    "score": 82,
    "post_edited_translation": "Ich entschuldige mich dafür, wir müssen die Erlaubnis einholen, um eine Bestellung mit Kontoinhaber zu besprechen. Ich entschuldige mich, falls dies zuvor geschehen kann, aber ohne die Erlaubnis des Kontoinhabers kann ich nicht in der Lage, dies mit Sie involvement.",
    "errors": {
        "critical": [],
        "major": [
            {"type": "accuracy/mistranslation", "source_error": "discuss", "target_error": "involvment", "correction": "besprechen", "short_desc": "'discuss' is mistranslated as 'involvment'"},
            {"type": "accuracy/omission", "source_error": "the account holder", "target_error": None, "correction": "Kontoinhaber", "short_desc": "'the account holder' is missing"}
        ],
\end{lstlisting}
\end{figure*}

\begin{figure*}[t]
\begin{lstlisting}[captionpos=b, label=lst:system-prompt]
        "minor": [
            {"type": "fluency/grammar", "source_error": None, "target_error": "wäre", "correction": "kann", "short_desc": "'wäre' is a bit awkward"},
            {"type": "fluency/register", "source_error": None, "target_error": "dir", "correction": "Sie", "short_desc": "'dir' should be 'Sie'"}
        ]
    },
}

Here is an example of a JSON input for Polish-Spanish translation:
{
    "source_language": "Polish",
    "source": "Szanowny Kliencie, informujemy, że Twoja przesyłka została dzisiaj nadana i powinna dotrzeć w ciągu 3 dni roboczych. Prosimy o sprawdzenie skrzynki mailowej w celu uzyskania numeru śledzenia.",
    "translation_language": "Spanish",
    "translation": "Estimado Cliente, informamos que tu envío fue hoy sobre y debe llegar en tres días trabajan. Por favor, chequea el mail para el número de rastrear."
}

And here is a corresponding JSON output with the score, post-edited translation and error annotations:
{
    "score": 45,
    "post_edited_translation": "Estimado Cliente, le informamos que su paquete fue enviado hoy y debería llegar en un plazo de 3 días hábiles. Por favor, revise su correo electrónico para obtener el número de seguimiento.",
    "errors": {
        "critical": [
            {"type": "accuracy/omission", "source_error": "przesyłka", "target_error": None, "correction": "paquete/envío", "short_desc": "Meaning is changed by omitting 'package' or 'shipment'"},
            {"type": "accuracy/mistranslation", "source_error": "nadana", "target_error": "sobre", "correction": "enviado", "short_desc": "'nadana' is incorrectly translated as 'sobre' (envelope/about)"}
        ],
        "major": [
            {"type": "fluency/grammar", "source_error": "roboczych", "target_error": "trabajan", "correction": "hábiles", "short_desc": "'trabajan' is incorrect word form, should be 'hábiles'"},
            {"type": "fluency/grammar", "source_error": "w ciągu", "target_error": "en", "correction": "en un plazo de", "short_desc": "Preposition 'en' or 'en un plazo de' is missing"},
            {"type": "style/awkwardness", "source_error": "sprawdzenie", "target_error": "chequea", "correction": "revise", "short_desc": "'chequea' is too informal, 'revise' is better for formal communication"}
        ],
        "minor": [
            {"type": "fluency/grammar", "source_error": "skrzynki mailowej", "target_error": "el mail", "correction": "el correo electrónico", "short_desc": "Missing article 'el' before 'mail'"},
            {"type": "fluency/grammar", "source_error": "śledzenia", "target_error": "rastrear", "correction": "seguimiento", "short_desc": "Gerund 'rastrear' is wrong, should be noun 'seguimiento'"}
        ]
    }
}

Here is an example of a JSON input for French-Italian translation:
{
    "source_language": "French",
    "source": "En raison d'une forte demande, il peut y avoir un léger retard dans le traitement de votre commande. Nous vous remercions de votre patience et de votre compréhension.",
    "translation_language": "Italian",
    "translation": "A causa di una forte domanda, potrebbe esserci un leggero ritardo nell'elaborazione del vostro ordine. Vi ringraziamo per la vostra pazienza e la vostra comprensione."
\end{lstlisting}
\end{figure*}

\begin{figure*}[h]
\begin{lstlisting}[caption={QE system prompt.}, captionpos=b, label=lst:system-prompt]
}

And here is a corresponding JSON output with the score, post-edited translation and error annotations:
{
    "score": 100,
    "post_edited_translation": "A causa di una forte domanda, potrebbe esserci un leggero ritardo nell'elaborazione del vostro ordine. Vi ringraziamo per la vostra pazienza e la vostra comprensione.",
    "errors": {
        "critical": [],
        "major": [],
        "minor": []
    }
}

If the whole translation is in the wrong language, return one "untranslated text" critical error for the whole sentence and no other errors.
IMPORTANT: All annotated errors and assigned ratings should refer only to the translation and not to the source text.

You will receive the {source_language}-{target_language} translation pair as a JSON object. Analyze the translation as discussed above and produce a JSON object with the analysis in response. Do not invent structural elements that are not present in the JSON examples above. The only allowed keys are "score", "post_edited_translation", "errors", "critical", "major", "minor", "type", "short_desc", "source_error", "target_error", "correction".
\end{lstlisting}
\end{figure*}

\begin{figure*}[h]
\begin{lstlisting}[caption={QE user prompt.}, captionpos=b, label=lst:user-prompt]
{
    "source_language": {source_language},
    "source": {source_text},
    "translation_language": {target_language},
    "translation": {target_text}
}
\end{lstlisting}
\end{figure*}

\FloatBarrier

\newpage

\section{Results}\label{appendix:results}

\FloatBarrier

\begin{table*}[t!h]
\centering
\begin{tabular}{|ccccc|cccc|}
\multicolumn{1}{c}{\rotatebox[origin=bl]{90}{COMET22}} & \rotatebox[origin=bl]{90}{COMETKiwi22} & \rotatebox[origin=bl]{90}{MetricX-25-QE} & \rotatebox[origin=bl]{90}{GEMBA-v2} & \multicolumn{1}{c}{\rotatebox[origin=bl]{90}{Gemini2\_5-Pro}} & \rotatebox[origin=bl]{90}{Eurollm-ESA} & \rotatebox[origin=bl]{90}{Qwen-ESA} & \rotatebox[origin=bl]{90}{Gemma-ESA} & \multicolumn{1}{c}{\rotatebox[origin=bl]{90}{Gemini\_3\_flash-ESA}} \\ \hline
\multicolumn{9}{|c|}{\cellcolor{gray!20}\textbf{\small Czech → German}} \\ \hline
0.77 & 0.61 & 0.72 & 0.84 & 0.90 & 0.76 & 0.80 & 0.80 & \textbf{0.91} \\ \hline
\multicolumn{9}{|c|}{\cellcolor{gray!20}\textbf{\small English → Italian}} \\ \hline
-    & 0.61 & 0.73 & 0.86 & 0.87 & 0.76 & \textbf{0.88} & 0.85 & 0.87 \\ \hline
\multicolumn{9}{|c|}{\cellcolor{gray!20}\textbf{\small English → Ukrainian}} \\ \hline
0.69 & 0.57 & 0.69 & 0.81 & 0.82 & 0.66 & 0.83 & 0.81 & \textbf{0.85} \\ \hline
\multicolumn{9}{|c|}{\cellcolor{gray!20}\textbf{\small Average}} \\ \hline
0.73 & 0.60 & 0.71 & 0.84 & 0.87 & 0.72 & 0.83 & 0.82 & \textbf{0.88} \\ \hline
\end{tabular}
\caption{System-level correlation between automated scores (both baselines and the proposed LLM-based QE metrics) and human evaluations for the WMT25 dataset, expressed as Soft Pairwise Accuracy. The table provides a comparative analysis for three specific language pairs and the average performance across all evaluated directions. In each row, the highest accuracy value is indicated in bold to highlight the most effective metric.}
\label{tab:accuracy_humeval}
\end{table*}

\begin{table*}[t!h]
    \centering
    \begin{tabular}{|ccccc|cccc|}
    \multicolumn{1}{c}{\rotatebox[origin=bl]{90}{COMET22}} & \rotatebox[origin=bl]{90}{COMETKiwi22} & \rotatebox[origin=bl]{90}{MetricX-25-QE} & \rotatebox[origin=bl]{90}{GEMBA-v2} & \multicolumn{1}{c}{\rotatebox[origin=bl]{90}{Gemini2\_5-Pro}} & \rotatebox[origin=bl]{90}{Eurollm-ESA} & \rotatebox[origin=bl]{90}{Qwen-ESA} & \rotatebox[origin=bl]{90}{Gemma-ESA} & \multicolumn{1}{c}{\rotatebox[origin=bl]{90}{Gemini\_3\_flash-ESA}} \\ \hline
    \multicolumn{9}{|c|}{\cellcolor{gray!20}\textbf{\small Czech → German}} \\ \hline
    0.57 & 0.55 & 0.58 & \textbf{0.63} & 0.61 & 0.39 & 0.49 & 0.48 & \textbf{0.63} \\ \hline
    \multicolumn{9}{|c|}{\cellcolor{gray!20}\textbf{\small English → Italian}} \\ \hline
    - & 0.60 & \textbf{0.64} & 0.63 & 0.61 & 0.42 & 0.44 & 0.42 & 0.61 \\ \hline
    \multicolumn{9}{|c|}{\cellcolor{gray!20}\textbf{\small English → Ukrainian}} \\ \hline
    0.58 & 0.61 & \textbf{0.63} & \textbf{0.63} & 0.59 & 0.46 & 0.41 & 0.43 & 0.59 \\ \hline
    \multicolumn{9}{|c|}{\cellcolor{gray!20}\textbf{\small Average}} \\ \hline
    0.57 & 0.59 & 0.62 & \textbf{0.63} & 0.60 & 0.42 & 0.45 & 0.45 & 0.61 \\ \hline
    \end{tabular}
    \caption{Segment-level correlation between automated scores (both baselines and the proposed LLM-based QE metrics) and human judgments on the WMT25 dataset. The evaluation utilizes the "group-by-item" accuracy with tie calibration to measure how well each metric aligns with human ratings at the segment level. The table details results for three language pairs along with their overall average. For each row, the maximum correlation value is highlighted in bold.}
    \label{tab:accuracy_humeval}
\end{table*}

\end{document}